\documentclass[10pt,twocolumn,letterpaper]{article}

\usepackage{iccv}      

\usepackage{multirow}

\definecolor{iccvblue}{rgb}{0.21,0.49,0.74}
\usepackage[pagebackref,breaklinks,colorlinks,allcolors=iccvblue]{hyperref}
\usepackage{footnotebackref}

\def\paperID{5406} 
\def\confName{ICCV}
\def\confYear{2025}

\title{Are Fast Methods Stable in Adversarially Robust Transfer Learning?}

\author{Joshua C. Zhao\\
Purdue University\\
{\tt\small zhao1207@purdue.edu}
\and
Saurabh Bagchi\\
Purdue University\\
{\tt\small sbagchi@purdue.edu}
}

\newcommand\blfootnote[1]{%
  \begingroup
  \renewcommand\thefootnote{}\footnote{#1}%
  \addtocounter{footnote}{-1}%
  \endgroup
}

\begin{document}
\maketitle
\begin{abstract}
Transfer learning is often used to decrease the computational cost of model training, as fine-tuning a model allows a downstream task to leverage the features learned from the pre-training dataset and quickly adapt them to a new task. This is particularly useful for achieving adversarial robustness, as adversarially training models from scratch is very computationally expensive. However, high robustness in transfer learning still requires adversarial training during the fine-tuning phase, which requires up to an order of magnitude more time than standard fine-tuning. In this work, we revisit the use of the fast gradient sign method (FGSM) in robust transfer learning to improve the computational cost of adversarial fine-tuning. We surprisingly find that FGSM is much more stable in adversarial fine-tuning than when training from scratch. In particular, FGSM fine-tuning does not suffer from any issues with catastrophic overfitting at standard perturbation budgets of $\varepsilon=4$ or $\varepsilon=8$. This stability is further enhanced with parameter-efficient fine-tuning methods, where FGSM remains stable even up to $\varepsilon=32$ for linear probing. We demonstrate how this stability translates into performance across multiple datasets. Compared to fine-tuning with the more commonly used method of projected gradient descent (PGD), on average, FGSM only loses 0.39\% and 1.39\% test robustness for $\varepsilon=4$ and $\varepsilon=8$ while using $4\times$ less training time. Surprisingly, FGSM may not only be a significantly more efficient alternative to PGD in adversarially robust transfer learning but also a well-performing one.
\end{abstract}    
\begin{figure}[t]
\vspace*{-1mm}
    \centering
   \includegraphics[width=0.95\linewidth,trim={0mm 0mm 0mm 0mm},clip]{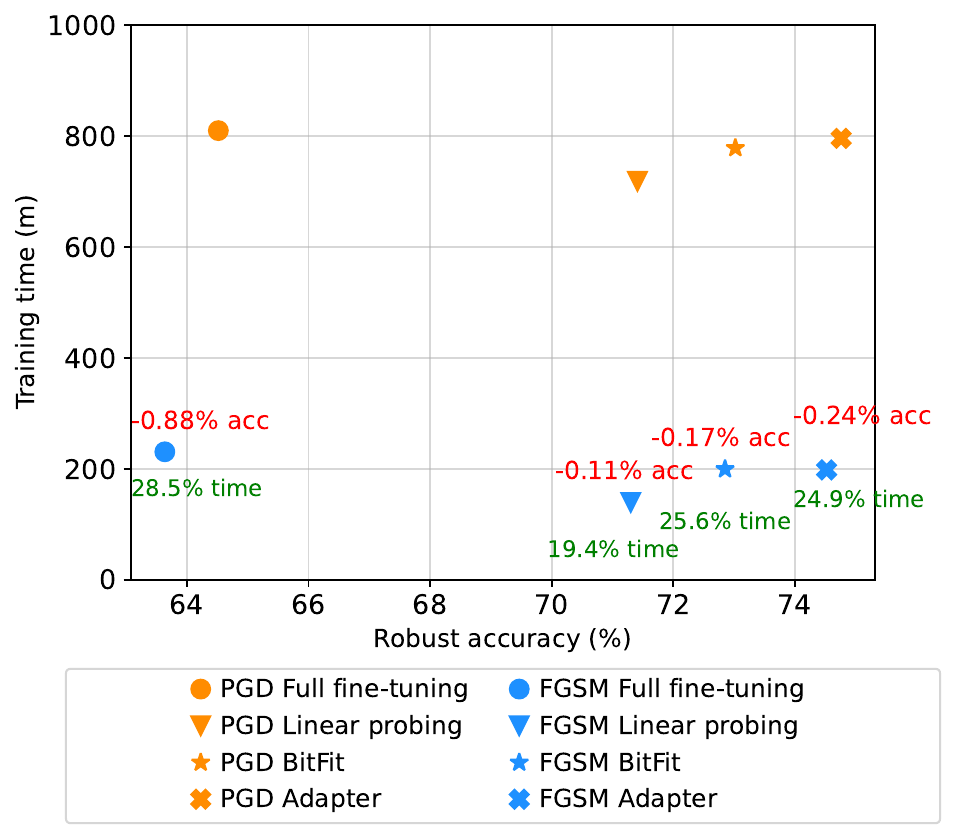}
  \caption{Adversarial fine-tuning using FGSM and PGD for 40 epochs on Caltech256 with a Swin transformer. The peak PGD-10 robustness and total training time (minutes) are reported. The red text shows the difference in robustness between FGSM and PGD, while the green text shows the percentage of PGD's training time used by FGSM. \textbf{On average, FGSM uses only 24.6\% of the time ($\boldsymbol{4.1\times}$ faster than PGD) and only loses 0.35\% robustness.}}
  \label{fig:intro} 
\vspace*{-5mm}
\end{figure}

\section{Introduction}
\label{sec:intro}

Deep learning has found incredible success in many areas, where applications in natural language processing~\cite{vaswani2017attention, devlin2019bert}, autonomous vehicles~\cite{van2018autonomous} or generative AI~\cite{rombach2022high, touvron2023llama} have become possible. However, with increasingly more models being applied everywhere, reliability becomes important. Within deep learning, one of the long-standing challenges has been adversarial examples, where small human-imperceptible changes to an input can fool a network prediction~\cite{szegedy2013intriguing}. Although many methods have been proposed to defend against adversarial examples~\cite{papernot2016distillation, guo2018countering}, many have also been broken over time~\cite{athalye2018obfuscated, carlini2017towards}. Of these, adversarial training (AT)~\cite{goodfellow2014explaining, madry2018towards} has stood the test of time to become the standard of providing model robustness. However, despite this, AT is still a very computationally expensive process, especially when training models from scratch.

Robust transfer learning~\cite{torrey2010transfer, weiss2016survey, salman2020adversarially} offers a good alternative. By fine-tuning a robust pre-trained network, this improves the cost of AT, with fewer training rounds required than training from scratch. Initial works exploring adversarial robustness in transfer learning completely avoided using AT during the fine-tuning process, showing that some robustness could still be maintained~\cite{Shafahi2020Adversarially}.
However, recent work has shown that achieving high downstream robustness still requires AT during the fine-tuning process (adversarial fine-tuning)~\cite{hua2024initialization, liu2023twins, xu2024autolora, hendrycks2019using}. Yet, while prior work has used adversarial fine-tuning to achieve high robustness, they have not considered the large computational cost added by adversarial fine-tuning. 

To illustrate this cost, we fine-tune a Swin transformer~\cite{liu2021swin} for 40 epochs on the Caltech256 dataset~\cite{griffin2007caltech}. On a NVIDIA A100 GPU, standard full (parameter) fine-tuning takes about 2.26 hours, while linear probing (tuning the classification head) takes 0.71 hours. When adversarially fine-tuning with a 7-step projected gradient descent attack (i.e., PGD~\cite{madry2018towards}, the standard method used for adversarial training or fine-tuning), full fine-tuning takes 13.50 hours, a $6.0\times$ increase in time compared to non-adversarial full fine-tuning. Adversarial linear probing requires 11.98 hours instead, a $16.9\times$ increase in time compared to non-adversarial linear probing. Although decreasing the number of fine-tuned parameters (going from full fine-tuning to linear probing) improves the computational cost, this improvement becomes more marginal compared to the overall cost of adversarial fine-tuning. This highlights a key point: \textit{Most of the cost of adversarial fine-tuning comes from the cost of generating adversarial examples, not updating the model parameters themselves.}


In this work, we investigate fast adversarial example generation methods for improving the time required for adversarial fine-tuning. In particular, we explore the use of the fast gradient sign method (FGSM)~\cite{goodfellow2014explaining} as an alternative to the commonly used PGD method. Although PGD uses multiple smaller projection steps during adversarial example generation (resulting in a stronger and more accurate estimate~\cite{madry2018towards}), FGSM opts for speed by using a single larger step instead. However, one of the main costs for FGSM's speed is catastrophic overfitting during AT~\cite{tramèr2018ensemble, kim2021understanding}, a well-known phenomenon in which a model trained with FGSM will lose all robustness to stronger attacks (such as PGD) once the model overfits during training~\cite{tramèr2018ensemble, kim2021understanding}. This is a core issue for FGSM AT when training from scratch, and multiple methods have been proposed to mitigate catastrophic overfitting~\cite{Wong2020Fast, andriushchenko2020understanding, wang2024preventing, park2021reliably}.

\blfootnote{Code will be provided.} 
However, we surprisingly discover that FGSM is much more stable in robust transfer learning than when training models from scratch. At standard $\ell_\infty$ perturbation budgets of $\varepsilon=4$ and $\varepsilon=8$, FGSM does not have any issue with catastrophic overfitting during adversarial fine-tuning. We further find that when FGSM is used in conjunction with parameter-efficient fine-tuning methods (PEFTs), the models become more stable. FGSM remains stable up to $\varepsilon=12$ for Adapter~\cite{houlsby2019parameter}, $\varepsilon=16$ for BitFit (bias)~\cite{ben-zaken-etal-2022-bitfit}, and even $\varepsilon=32$ with linear probing, all of which are much higher than the perturbation budget typically required for AT.

This stability in adversarial fine-tuning creates an alternative setting for FGSM where avoiding catastrophic overfitting is not an important concern. Compared to random initialization~\cite{Wong2020Fast} or gradient alignment (GradAlign)~\cite{andriushchenko2020understanding}, regularization methods that FGSM requires when training models from scratch, the standard FGSM attack without any additional regularization performs just as well in adversarial fine-tuning (even using 33.8\% less time than GradAlign). Compared to PGD fine-tuning, FGSM also offers comparable performance at a much lower cost. For $\varepsilon=4$ and $\varepsilon=8$ on a Swin transformer, FGSM uses $4\times$ less training time while having only 0.39\% and 1.39\% lower robustness on average (and having 0.35\% and 1.21\% higher natural accuracy at the same time).




Our main contributions are as follows:
\begin{itemize}
    \item We demonstrate FGSM's stability during adversarial fine-tuning. Compared to training from scratch, FGSM remains stable and avoids catastrophic overfitting at the standard values of $\varepsilon=4$ and $\varepsilon=8$. This stability is further improved with parameter-efficient fine-tuning methods. With linear probing, FGSM can remain stable even up to $\varepsilon$ as large as 32.
    \item We compare different FGSM methods and show that additional regularization is not required for stability. Although useful when training from scratch, these methods offer no performance improvement for adversarial fine-tuning while potentially increasing training time. We further show that the main difference between PGD and FGSM in very large $\varepsilon$ comes from the attack strength.
    \item We demonstrate that FGSM offers performance comparable to PGD at a much lower cost in robust transfer learning. When adversarially fine-tuning a Swin transformer, FGSM uses $4\times$ less time with only $0.39\%$ and 1.39\% lower robustness on average (while also having $0.35\%$ and 1.21\% higher natural accuracy on average) across all datasets for $\varepsilon=4$ and $\varepsilon=8$, respectively.
\end{itemize}

\section{Background and related work}

\textbf{Adversarial robustness.} Deep learning models are susceptible to adversarial examples, where a small perturbation away from a natural example can cause networks to misclassify otherwise correct examples~\cite{szegedy2013intriguing}. One of the most commons ways for generating adversarial examples is through projected gradient descent (PGD)~\cite{madry2018towards}:
\begin{equation}\label{eq:PGD}
    x_{adv}^{t+1} = \mathcal{P}(x_{adv}^t + \alpha \text{sign}(\nabla_{x_{adv}^{t}}L(\theta,x_{adv}^{t},y)))
\end{equation}
PGD is an iterative attack, where the adversarial example $x_{adv}=x+\delta$ is generated using multiple smaller projection steps. In order to maintain ``imperceptibility," these adversarial examples are bounded by $\lVert{\delta}\rVert_p \leq \varepsilon$ (we use an $\ell_\infty$ budget in this paper, which means that the perturbation of any pixel must be $\leq\frac{\varepsilon}{255}$), where $\varepsilon$ is a small value, typically 4 or 8 for image datasets. An alternative to PGD is the fast gradient sign method (FGSM)~\cite{goodfellow2014explaining}, a faster method where only a single larger step is taken.

One of the most well-accepted methods for improving model robustness is adversarial training (AT):
\begin{equation}\label{eq:AT_formulation}
    \min_{\theta}\mathbb{E}_{(x,y)\sim D} \left[ \max_{\delta\in \bigtriangleup}\mathcal{L}(\theta,x+\delta,y) \right]
\end{equation}
In AT, the inner maximization aims to generate examples that maximize the model loss (adversarial examples), and the outer minimization is a minimization of the model loss on these examples. The standard method for inner maximization is PGD~\cite{madry2018towards}, and although FGSM can be used instead~\cite{goodfellow2014explaining}, this results in the phenomenon of catastrophic overfitting, where a model trained from scratch with FGSM would overfit and lose almost all the robustness to stronger attacks~\cite{tramèr2018ensemble, Wong2020Fast, kim2021understanding}. This occurs across model architectures~\cite{wu2022towards, shao2021adversarial} and although early stopping could help mitigate the effects~\cite{andriushchenko2020understanding}, the model performance was often still worse than that of PGD AT~\cite{madry2018towards, zhang2019theoretically}.

Multiple works have been proposed to avoid catastrophic overfitting. For example, \cite{shafahi2019adversarial} initialized adversarial examples using the perturbation from a previous batch. In a subsequent work, \cite{Wong2020Fast} found that FGSM + random initialization (FGSM + RI) allowed FGSM to train models from scratch that were still robust to stronger attacks like PGD. Another work showed how FGSM was only stable for lower perturbation budgets and improved this through a gradient alignment regularizer called GradAlign~\cite{andriushchenko2020understanding}.

\textbf{Robust transfer learning.} Robust transfer learning~\cite{torrey2010transfer, weiss2016survey, salman2020adversarially} offers an efficient alternative to AT from scratch, as fine-tuning a model is more efficient in terms of training rounds. Fine-tuning a pre-trained model can even help alleviate some sample complexity problems~\cite{schmidt2018adversarially, carmon2019unlabeled, wang2023better} and improve performance, since pre-training datasets are often much larger than downstream datasets. Initial works avoided using any AT for robust transfer learning and instead focused on preserving robustness~\cite{Shafahi2020Adversarially}. However, achieving high robustness still requires adversarial fine-tuning (adversarial training during fine-tuning)~\cite{hua2024initialization, liu2023twins, hendrycks2019using}. Several methods have been proposed with adversarial fine-tuning. For example, TWINS~\cite{liu2023twins} splits the network into two ``twin" models during adversarial fine-tuning, where one of the networks preserves the pre-training batch norm statistics by freezing them. Another work, AutoLoRa~\cite{xu2024autolora}, applies natural fine-tuning to the LoRA branch and adversarial fine-tuning to the feature extractor. 

Robust linear initialization (RoLI) proposes an additional robust initialization to the linear classification head prior to adversarial fine-tuning and explores this in conjunction with parameter-efficient fine-tuning methods (PEFTs)~\cite{hua2024initialization}. In addition to full (parameter) fine-tuning and linear probing (tuning of the classification head), many other PEFTs have been developed such as BitFit~\cite{ben-zaken-etal-2022-bitfit} which fine-tunes the biases, Adapters~\cite{houlsby2019parameter} which introduces additional bottleneck layers, or LoRA~\cite{hu2021lora} which applies an additional tuning branch to the query, key, and value weights. These PEFTs only fine-tune their specific parameters along with the final classification layer, which decreases the number of tuned parameters and reduces the memory overhead. Given the smaller number of tuned parameters, PEFTs can also decrease the fine-tuning time. However, they do not decrease the adversarial example generation time, which makes up the bulk of time in adversarial fine-tuning.

In this work, we explore efficient adversarially robust transfer learning using FGSM (with and without PEFTs), demonstrating the surprising stability and performance of FGSM during adversarial fine-tuning.
\section{FGSM stability in adversarial fine-tuning}
\label{sec:fgsm_vary_eps}

Previous works have explored the use of FGSM for efficient AT from scratch~\cite{Wong2020Fast, andriushchenko2020understanding, goodfellow2014explaining} and have shown that catastrophic overfitting is a key problem~\cite{tramèr2018ensemble, kim2021understanding}. The sudden drop in test robustness against stronger attacks like PGD is a major issue that forces FGSM to rely on early stopping, random initialization, or other regularization. However, this has not been explored in robust transfer learning. Hence, we first ask: \textit{Is the stability of FGSM any different during adversarial fine-tuning?}

\begin{figure}[t]
\vspace*{-0mm}
  \centering
  \begin{subfigure}{1.0\linewidth}
    \includegraphics[width=0.9\linewidth,trim={0mm 0mm 0mm 0mm},clip]{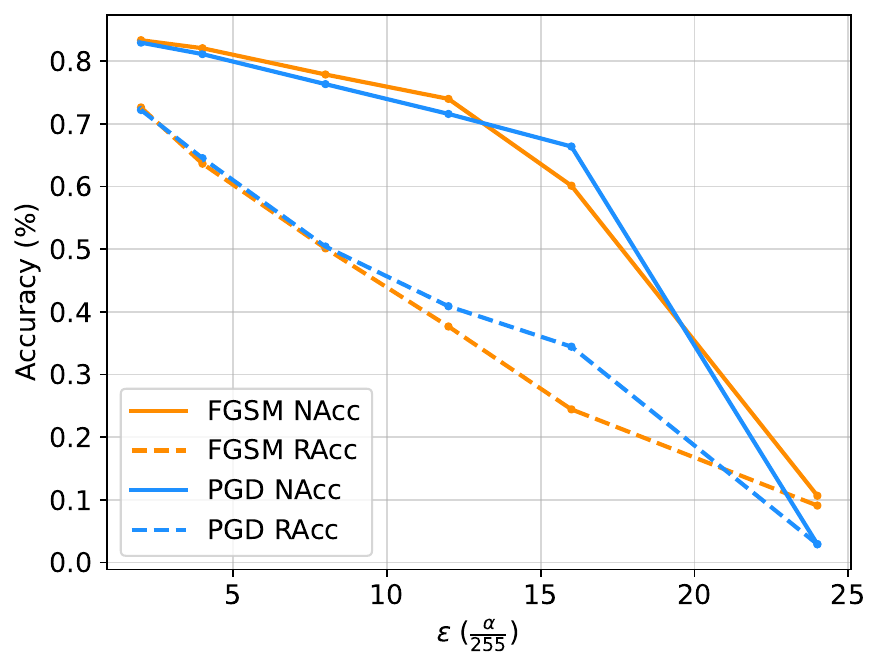}
    \caption{Full fine-tuning}
    \label{fig:fgsm_pgd_FFT_veps}
  \end{subfigure}
  \vspace*{3mm}
    \begin{subfigure}{1.0\linewidth}
    \includegraphics[width=0.9\linewidth,trim={0mm 0mm 0mm 0mm},clip]{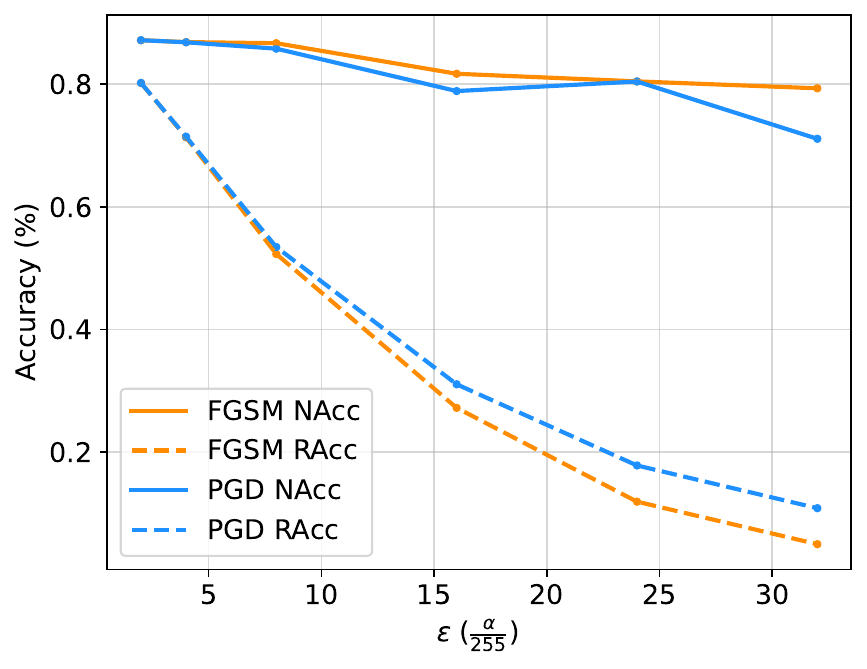}
    \caption{Linear probing}
    \label{fig:fgsm_pgd_LP_veps}
  \end{subfigure}
  \vspace*{-10mm}
  \caption{\label{fig:vary-epsilon} FGSM vs. PGD (a) full fine-tuning and (b) linear probing natural accuracy (NAcc) and robust accuracy (RAcc) for varying $\varepsilon$ values on Caltech256 with a Swin transformer. At $\varepsilon=16$, FGSM full fine-tuning incurs a larger performance drop compared to PGD although both fail at $\varepsilon=24$. Linear probing is stable for both FGSM and PGD even with large $\varepsilon$ values.}
  \vspace*{-5mm}
\end{figure}

\subsection{Fine-tuning with PGD and FGSM}
\label{sec:sec3_stability}

\begin{figure}[t]
\vspace*{-0mm}
    \centering
   \includegraphics[width=0.9\linewidth,trim={0mm 0mm 0mm 0mm},clip]{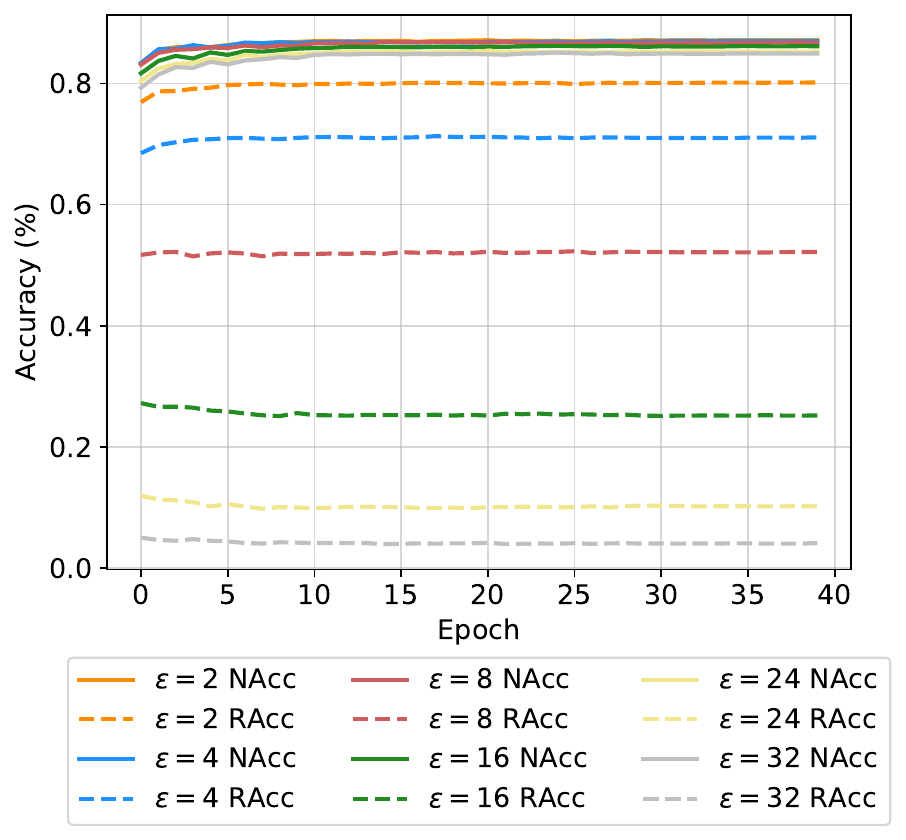}
  \caption{Natural accuracy (solid line) and robustness (dashed line) for FGSM linear probing when trained with different $\varepsilon$. The natural accuracy and robust accuracy of the model are stable regardless of the size of $\varepsilon$, without catastrophic overfitting.}
  \label{fig:fgsm_LP_vary_eps} 
\vspace*{-5mm}
\end{figure}

To understand more about the stability of FGSM, we begin by comparing FGSM and PGD under varying $\varepsilon$ budgets for adversarial fine-tuning. We use a 7-step PGD attack (PGD-7) with a step size of $\frac{\varepsilon}{4}$~\cite{madry2018towards} and a standard FGSM attack with a single step of size $\alpha=\varepsilon$~\cite{goodfellow2014explaining}. We use a Swin-B transformer~\cite{liu2021swin} with pre-trained weights downloaded from the robustly trained models of ARES 2.0~\cite{dong2020benchmarking}.

We first consider two scenarios for fine-tuning: full (parameter) fine-tuning and linear probing (fixed feature fine-tuning where only the classification head is tuned). Figures~\ref{fig:fgsm_pgd_FFT_veps} and \ref{fig:fgsm_pgd_LP_veps} show the peak PGD-10 test robustness for full fine-tuning and linear probing, respectively, across multiple different $\varepsilon$ when fine-tuning for 40 epochs on Caltech256. Detailed numbers are reported in the supplementary Section~\ref{sec:additional_results}. For full fine-tuning with $\varepsilon=2,4,8,12$, FGSM has -0.4\%, 0.9\%, 0.3\%, and 3.2\% lower peak robustness than PGD. At $\varepsilon=16$, catastrophic overfitting becomes a problem for FGSM, and the gap increases to 6.3\% and 10.0\% lower natural accuracy and robustness (catastrophic overfitting is shown in the FGSM testing accuracy and robustness curves in the supplementary Figure~\ref{fig:fgsm_fft_vary_eps}). At $\varepsilon=24$, neither FGSM nor PGD converges to a good solution.

Linear probing observes a consistent trend with the robustness gap slowly increasing with larger $\varepsilon$. For $\varepsilon=2,4,8,16,24,32$, the robustness difference between FGSM and PGD is 0.0\%, 0.1\%, 1.1\%, 3.8\%, 5.9\%, and 5.9\%. Similar to full fine-tuning, FGSM achieves close performance until $\varepsilon\geq16$. However, in contrast to full fine-tuning, \textit{FGSM linear probing never catastrophically overfits}. Figure~\ref{fig:fgsm_LP_vary_eps} shows the test accuracy (solid lines) and robustness (dashed lines) curves over the 40 fine-tuning epochs for each FGSM run. For all $\varepsilon$, FGSM converges to a solution without catastrophic overfitting, and the natural accuracy and robustness remain consistent during training. This even holds for $\varepsilon$ as large as $32$.

Although FGSM catastrophic overfitting still occurs during full fine-tuning, it is more stable. When training with FGSM from scratch, prior work has found that even $\varepsilon=7$ can cause models to have problems with catastrophic overfitting~\cite{andriushchenko2020understanding}. In the supplementary Figure~\ref{fig:fgsm_pgd_c256_scratch}, we show how a ResNet-50 trained from scratch with $\varepsilon=8$ has a near-zero PGD robustness at the end.\footnote{Given the high cost of adversarially training transformer models from scratch, we train a ResNet-50. Other works have also demonstrated catastrophic overfitting for transformers when training from scratch~\cite{wu2022towards, shao2021adversarial, gopal2025safer}.} However, FGSM full fine-tuning is very stable at $\varepsilon=8$ and only begins to overfit at $\varepsilon=12$, as shown in the supplementary Figure~\ref{fig:fgsm_fft_vary_eps}. 

This stability of FGSM is further enhanced by linear probing, with no catastrophic overfitting up to $\varepsilon=32$. We hypothesize that this comes from the feature-restricting nature of PEFTs~\cite{hua2024initialization, salman2020adversarially, houlsby2019parameter}. By constraining the number of fine-tuned parameters, this also restricts how much the model features can be altered. This property also functions as a form of regularization to prevent the model from converging to the ``degenerate minimum"~\cite{tramèr2018ensemble} that causes catastrophic overfitting. In the supplementary Section~\ref{sec:additional_results}, we apply the same experiment with the BitFit and Adapter PEFTs. Similarly to linear probing, both methods add additional stability to fine-tuning, where BitFit remains stable through $\varepsilon=16$ and Adapter remains stable through $\varepsilon=12$.

Overall, FGSM is surprisingly stable for adversarial fine-tuning. Through the initialization of robust pre-trained model parameters, full fine-tuning remains stable when $\varepsilon=8$. Although this is enough, as $\varepsilon>8$ is usually not used for AT, we also find that PEFTs further increase the stability of FGSM, even up to $\varepsilon=32$ for linear probing. In the next section, we compare several FGSM variants in adversarial fine-tuning to further show the impact of stability. 

\subsection{FGSM variants in robust transfer learning}
\label{sec:FGSM_variants}

\begin{table}[t]
\begin{center}
\begin{tabular}{l|l|ccc}
\hline
                                                                           & Method           & NAcc                               & RAcc                               & \begin{tabular}[c]{@{}c@{}}Time\\ (min)\end{tabular} \\ \hline
\multirow{5}{*}{FGSM}                                                      & Full param       & 77.87                              & 50.13                              & 230.7                                                \\
                                                                           & Linear probe     & 86.74                              & 52.21                              & 139.3                                                \\
                                                                           & BitFit (bias)    & 84.81                              & 55.50                              & 199.3                                                \\
                                                                           & Adapter          & 86.92                              & 56.86                              & 198.0                                                \\ \cline{2-5} 
                                                                           & \textbf{Average} & \multicolumn{1}{l}{\textbf{84.09}} & \multicolumn{1}{l}{\textbf{53.68}} & \multicolumn{1}{l}{\textbf{191.8}}                   \\ \hline
\multirow{5}{*}{FGSM+RI}                                                   & Full param       & 78.63                              & 49.36                              &      231.2                                                \\
                                                                           & Linear probe     & 85.43                              & 52.07                              &      139.6                                                \\
                                                                           & BitFit (bias)    & 85.38                              & 55.21                              &      199.3                                                \\
                                                                           & Adapter          & 87.63                              & 56.47                              &    198.2                                                  \\ \cline{2-5} 
                                                                           & \textbf{Average} & \multicolumn{1}{l}{\textbf{84.27}} & \multicolumn{1}{l}{\textbf{53.28}} & \multicolumn{1}{l}{\textbf{192.1}}                        \\ \hline
\multirow{5}{*}{\begin{tabular}[c]{@{}l@{}}FGSM\\ +GradAlign\end{tabular}} & Full param       &     77.20                               &   49.64                                 &     328.3                                                 \\
                                                                           & Linear probe     &         85.32                           &    52.40                                &      236.5                                                \\
                                                                           & BitFit (bias)    &        84.73                            &    55.80                                &    296.2                                                  \\
                                                                           & Adapter          &     86.80                               &    56.88                                &    298.2                                                  \\ \cline{2-5} 
                                                                           & \textbf{Average} & \multicolumn{1}{l}{\textbf{83.51}}      & \multicolumn{1}{l}{\textbf{53.68}}      & \multicolumn{1}{l}{\textbf{289.8}}                        \\ \hline
\end{tabular}
\end{center}
\vspace*{-3mm}
\caption{\label{tab:fgsm_variants_c256_eps8} Fine-tuning accuracy and time for FGSM, FGSM + random initialization (FGSM+RI)~\cite{Wong2020Fast}, and FGSM+GradAlign~\cite{andriushchenko2020understanding} on Caltech256 with $\varepsilon=8$. All three methods achieve very similar performance (with FGSM+RI having marginally lower robustness and  marginally higher natural accuracy. FGSM+GradAlign achieves the same average robustness as standard FGSM while requiring 51.1\% more time.}
\vspace*{-3mm}
\end{table}

Several methods have been proposed for fast adversarial training, with two notable ones including FGSM with random initialization (FGSM+RI)~\cite{Wong2020Fast} and FGSM with gradient alignment (FGSM+GradAlign)~\cite{andriushchenko2020understanding}. Both methods allow FGSM AT models to achieve PGD robustness when training from scratch. FGSM+RI uses a larger FGSM step of $\alpha=1.25\varepsilon$ (adversarial perturbations are still clipped to $\{-\varepsilon,\varepsilon\}$) while adding an random initialization step of $\delta=Uniform(-\varepsilon,\varepsilon)$. FGSM+GradAlign adds an additional regularizer term in the form of $1-Cos(\nabla_x(x,y,\theta),\nabla_x(x+\delta))$. This term minimizes the cosine distance between the gradient of the image $x$ at the points $x$ and $x+\delta$ (where $\delta=Uniform(-\varepsilon,\varepsilon)$), helping to align the gradient of FGSM within the perturbation ball.

Table~\ref{tab:fgsm_variants_c256_eps8} reports the performance and training time of each method on Caltech256 with $\varepsilon=8$. Surprisingly, there is little difference in natural accuracy or robustness between each method. On average, standard FGSM and FGSM+GradAlign achieve the same robustness, whereas FGSM+RI is slightly lower. However, FGSM+GradAlign also uses an additional 51.1\% time compared to standard FGSM due to the additional gradient computation in the regularizer. Overall, the simple FGSM attack without regularization performs well in adversarial fine-tuning. This is in stark contrast to training from scratch, where FGSM+RI can result in models with $>7\%$ higher robustness than standard FGSM~\cite{Wong2020Fast} and FGSM+GradAlign results in models with $>10\%$ higher robustness~\cite{andriushchenko2020understanding}.

This again highlights the stability for FGSM in robust transfer learning. Although these regularization methods are necessary when training models from scratch, they are not necessary for adversarial fine-tuning. FGSM remains stable at $\varepsilon=8$ for all fine-tuning methods and, as a result, additional regularization aimed at preventing catastrophic overfitting offers no performance benefits. In some cases, the methods even increase the computational cost instead.

\subsection{FGSM and PGD estimation difference}
\label{sec:cosine}

\begin{figure}[t]
\vspace*{-0mm}
    \centering
   \includegraphics[width=0.9\linewidth,trim={0mm 0mm 0mm 0mm},clip]{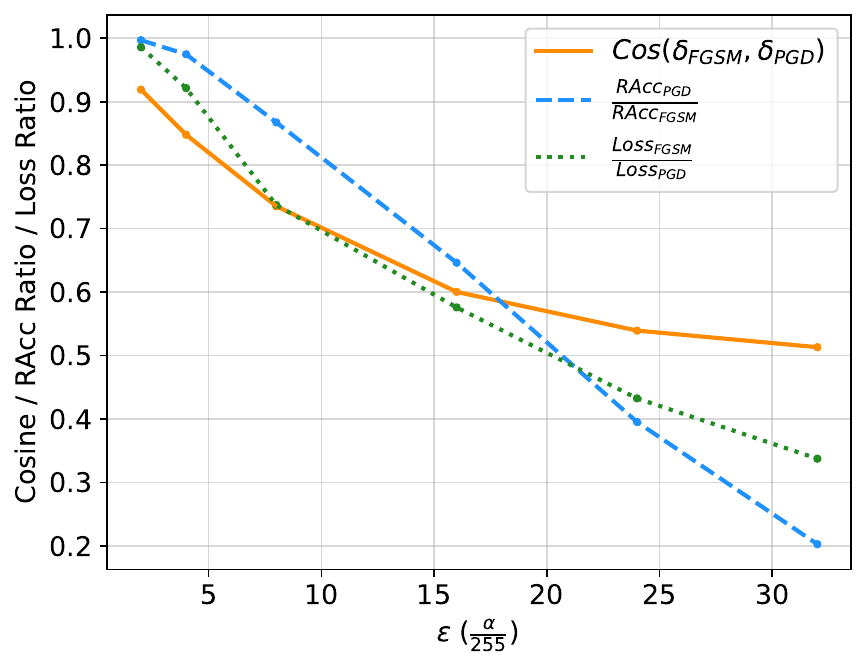}
   
  \caption{Average cosine distance between FGSM and PGD adversarial examples generated on the Caltech256 test set for different $\varepsilon$. The average loss ratio ($\frac{Loss_{FGSM}}{Loss_{PGD}}$) and robust accuracy ($\frac{RAcc_{PGD}}{RAcc_{FGSM}}$) ratios are also shown. As $\varepsilon$ becomes larger, the difference between FGSM and PGD also becomes larger.}
  \label{fig:fgsm_pgd_cosine_distance} 
\vspace*{-5mm}
\end{figure}

\begin{table*}[ht]
\begin{center}
\begin{tabular}{ll|l|cccccccccc}
\hline
                                              & \multicolumn{1}{c|}{}                   & \multicolumn{1}{c|}{}                                                                               & \multicolumn{2}{c|}{Caltech256}                                               & \multicolumn{2}{c|}{Dogs}                                                     & \multicolumn{2}{c|}{Flowers}                                                  & \multicolumn{2}{c|}{CIFAR-10}                                                 & \multicolumn{2}{c}{CIFAR-100}                                                 \\
\multirow{-2}{*}{}                            & \multicolumn{1}{c|}{\multirow{-2}{*}{}} & \multicolumn{1}{c|}{\multirow{-2}{*}{\begin{tabular}[c]{@{}c@{}}Fine-tuning\\ method\end{tabular}}} & NAcc                                  & \multicolumn{1}{c|}{RAcc}             & NAcc                                  & \multicolumn{1}{c|}{RAcc}             & NAcc                                  & \multicolumn{1}{c|}{RAcc}             & NAcc                                  & \multicolumn{1}{c|}{RAcc}             & NAcc                                  & RAcc                                  \\ \hline
\multicolumn{1}{l|}{}                         &                                         & Full param                                                                                          & 76.33                                 & 50.24                                 & 65.28                                 & 24.48                                 & 80.06                                 & 48.43                                 & 91.51                                 & 60.08                                 & 72.35                                 & 36.17                                 \\
\multicolumn{1}{l|}{}                         &                                         & Linear probe                                                                                        & 85.76                                 & 53.43                                 & 88.02                                 & 34.15                                 & 80.66                                 & 41.63                                 & 85.51                                 & 35.19                                 & 68.15                                 & 24.75                                 \\
\multicolumn{1}{l|}{}                         &                                         & BitFit (bias)                                                                                       & 83.53                                 & 57.42                                 & 86.19                                 & 37.18                                 & 84.11                                 & 49.70                                 & 89.96                                 & 52.71                                 & 72.33                                 & 35.28                                 \\
\multicolumn{1}{l|}{}                         & \multirow{-4}{*}{PGD}                   & Adapter                                                                                             & 83.18                                 & 57.78                                 & 84.95                                 & 36.77                                 & 82.96                                 & 49.85                                 & 92.07                                 & 58.88                                 & 75.09                                 & 36.90                                 \\ \cline{2-13} 
\multicolumn{1}{l|}{}                         &                                         & Full param                                                                                          & 77.87                                 & 50.13                                 & 66.75                                 & 23.82                                 & 80.87                                 & 47.37                                 & 92.38                                 & 57.60                                  & 72.39                                 & 35.74                                 \\
\multicolumn{1}{l|}{}                         &                                         & Linear probe                                                                                        & 86.74                                 & 52.21                                 & 89.11                                 & 33.37                                 & 80.53                                 & 40.38                                 & 87.31                                 & 32.46                                 & 68.78                                 & 23.42                                 \\
\multicolumn{1}{l|}{}                         &                                         & BitFit (bias)                                                                                       & 84.81                                 & 55.50                                  & 86.49                                 & 35.86                                 & 84.60                                 & 48.76                                 & 91.84                                 & 52.29                                 & 75.42                                 & 33.01                                 \\
\multicolumn{1}{l|}{}                         &                                         & Adapter                                                                                             & 86.92                                 & 56.86                                 & 86.76                                 & 34.95                                 & 81.54                                 & 48.71                                 & 94.03                                 & 56.21                                 & 77.02                                 & 34.55                                 \\ \cline{3-13} 
\multicolumn{1}{l|}{\multirow{-9}{*}{Swin}} & \multirow{-5}{*}{FGSM}                  & \textbf{Average diff.}                                                                              & {\color[HTML]{32CB00} \textbf{+1.89}} & {\color[HTML]{CB0000} \textbf{-1.04}} & {\color[HTML]{32CB00} \textbf{+1.17}} & {\color[HTML]{CB0000} \textbf{-1.15}} & {\color[HTML]{CB0000} \textbf{-0.06}} & {\color[HTML]{CB0000} \textbf{-1.10}} & {\color[HTML]{32CB00} \textbf{+1.63}} & {\color[HTML]{CB0000} \textbf{-2.08}} & {\color[HTML]{32CB00} \textbf{+1.42}} & {\color[HTML]{CB0000} \textbf{-1.60}} \\ \hline
\multicolumn{1}{l|}{}                         &                                         & Full param                                                                                          & 74.80                                  & 46.15                                 & 62.11                                 & 20.85                                 & 81.61                                 & 49.83                                 & 90.89                                 & 59.42                                 & 69.92                                 & 35.67                                 \\
\multicolumn{1}{l|}{}                         &                                         & Linear probe                                                                                        & 83.21                                 & 46.36                                 & 81.93                                 & 23.54                                 & 80.24                                 & 37.03                                 & 84.17                                 & 35.58                                 & 66.66                                 & 24.29                                 \\
\multicolumn{1}{l|}{}                         &                                         & BitFit (bias)                                                                                       & 81.44                                 & 51.99                                 & 77.96                                 & 28.28                                 & 80.91                                 & 46.14                                 & 87.84                                 & 51.35                                 & 71.02                                 & 33.14                                 \\
\multicolumn{1}{l|}{}                         & \multirow{-4}{*}{PGD}                   & Adapter                                                                                             & 81.80                                 & 52.60                                 & 72.41                                 & 26.67                                 & 82.00                                 & 46.82                                 & 91.89                                 & 58.22                                 & 74.02                                 & 36.29                                 \\ \cline{2-13} 
\multicolumn{1}{l|}{}                         &                                         & Full param                                                                                          & 74.68                                 & 45.07                                 & 64.04                                 & 18.47                                 & 81.98                                 & 49.13                                 & 92.07                                 & 57.09                                 & 71.69                                 & 33.84                                 \\
\multicolumn{1}{l|}{}                         &                                         & Linear probe                                                                                        & 83.78                                 & 44.77                                 & 82.90                                 & 22.51                                 & 80.37                                 & 34.51                                 & 86.45                                 & 32.41                                 & 68.01                                 & 22.42                                 \\
\multicolumn{1}{l|}{}                         &                                         & BitFit (bias)                                                                                       & 81.64                                 & 50.51                                 & 78.26                                 & 26.48                                 & 81.18                                 & 43.94                                 & 89.21                                 & 49.62                                 & 72.30                                  & 31.33                                 \\
\multicolumn{1}{l|}{}                         &                                         & Adapter                                                                                             & 83.21                                 & 50.90                                 & 78.85                                 & 25.49                                 & 82.37                                 & 45.34                                 & 93.15                                 & 55.42                                 & 76.07                                 & 33.96                                 \\ \cline{3-13} 
\multicolumn{1}{l|}{\multirow{-9}{*}{ViT}}  & \multirow{-5}{*}{FGSM}                  & \textbf{Average diff.}                                                                              & {\color[HTML]{32CB00} \textbf{+0.52}} & {\color[HTML]{CB0000} \textbf{-1.46}} & {\color[HTML]{32CB00} \textbf{+2.41}} & {\color[HTML]{CB0000} \textbf{-1.60}} & {\color[HTML]{32CB00} \textbf{+0.29}} & {\color[HTML]{CB0000} \textbf{-1.73}} & {\color[HTML]{32CB00} \textbf{+1.53}} & {\color[HTML]{CB0000} \textbf{-2.51}} & {\color[HTML]{32CB00} \textbf{+1.61}} & {\color[HTML]{CB0000} \textbf{-1.96}} \\ \hline
\end{tabular}
\end{center}
\vspace*{-3mm}
\caption{\label{tab:fgsm_pgd_performance_eps8}Performance of FGSM and PGD for $\varepsilon=8$ when fine-tuning Swin and ViT transformers for multiple datasets. Individual performance is given for each fine-tuning method, dataset, and model combination. The difference between FGSM and PGD averaged across fine-tuning methods is also reported for each dataset. Across all datasets combined, FGSM averages 1.39\% lower robustness and 1.21\% higher natural accuracy for the Swin transformer (1.85\% and 1.27\% for ViT).}
\vspace*{-0mm}
\end{table*}

Although FGSM is more stable (especially with PEFTs), there is still an apparent performance gap between PGD and FGSM for large $\varepsilon$. Even though models are almost never trained with $\varepsilon$ as large as 24 or 32, it is still important to understand where the performance gap comes from. 

Using linear probing as an example, while the FGSM models are still stable at $\varepsilon=32$, the robustness of the model is only $\frac{1}{2}$ that of the PGD model (although both are still low). Since catastrophic overfitting does not occur, this difference can instead be attributed to attack strength, as taking multiple smaller steps will lead to a more accurate estimation of an adversarial perturbation compared to a single large step~\cite{madry2018towards}. This is still the case for FGSM and PGD in adversarial fine-tuning.


To visualize this difference, we compare the perturbations of adversarial examples, $\delta$, generated by FGSM and PGD-7 for different $\varepsilon$ values. We compute this using a cosine similarity:
\vspace*{-1mm}
\begin{equation}
    Cos(\delta_{FGSM}(x,y,\theta), \delta_{PGD-7}(x,y,\theta))
\vspace*{-1mm}
\end{equation} 
with $\delta(x,y,\theta)=X_{adv}-X$ and a $Cos$ similarity of 1 indicating that the examples are the same and -1 indicating that they are the opposite. The use of cosine similarity is similar to GradAlign~\cite{andriushchenko2020understanding}, although we use cosine similarity to quantify how adversarial examples of FGSM and PGD differ with larger $\varepsilon$, and GradAlign uses this to regularize FGSM and visualize catastrophic overfitting.

Figure~\ref{fig:fgsm_pgd_cosine_distance} shows the average cosine similarity between adversarial examples generated using FGSM and PGD-7 on the Caltech256 test set for different $\varepsilon$. For each $\varepsilon$, we use the FGSM linear probing model in Section~\ref{sec:sec3_stability} that achieves the highest robustness. We also plot the loss ratio ($\frac{Loss_{FGSM}}{Loss_{PGD}}$) and the robust accuracy ratio ($\frac{RAcc_{PGD}}{RAcc_{FGSM}}$) of the adversarial examples, where a higher ratio indicates greater similarity in attack strength. From Figure~\ref{fig:fgsm_pgd_cosine_distance}, we see that as $\varepsilon$ becomes larger, the cosine similarity between the perturbations of $\delta_{FGSM}$ and $\delta_{PGD}$ decreases. At the same time, both the loss ratio and the robust accuracy ratio also decrease. These indicate that, with larger $\varepsilon$, the FGSM attack becomes more dissimilar and also weaker relative to PGD.

These differences only become a problem with large $\varepsilon$. Adversarial training and fine-tuning is generally performed with smaller values of $\varepsilon=4$ or $\varepsilon=8$. Here, the similarity between adversarial examples generated by FGSM and PGD is still high, allowing FGSM to serve as a fast and good alternative. Next, we compare the performance of FGSM and PGD adversarial fine-tuning on multiple datasets using the Swin and ViT transformer models.
\section{Experiments}
\label{sec:experiments}

In this section, we compare the performance of FGSM and PGD adversarial fine-tuning in terms of accuracy and computational cost using standard $\varepsilon$ values of 4 and 8. For FGSM fine-tuning, we use a standard attack without any additional regularization or random initialization. For PGD, we use a 7-step attack (PGD-7). All experiments are run using a NVIDIA A100 GPU, and for the pre-trained model, we use the model weights of the Swin-B~\cite{liu2021swin} and ViT-B/16~\cite{dosovitskiy2021an} transformer models robustly trained on ImageNet-1K~\cite{deng2009imagenet} from ARES 2.0~\cite{dong2020benchmarking}. We fine-tune on the Caltech256~\cite{griffin2007caltech}, Stanford Dogs~\cite{khosla2011novel}, Oxford Flowers102~\cite{nilsback2008automated}, CIFAR-10 and CIFAR-100~\cite{krizhevsky2009learning} datasets. For fine-tuning methods, we compare full (parameter) fine-tuning, linear probing (final classification layer), BitFit (bias)~\cite{ben-zaken-etal-2022-bitfit}, and Adapter~\cite{houlsby2019parameter}. More training details and some additional results are given in the supplementary Section~\ref{sec:suppl_exp_details}.

\begin{table*}[ht]
\begin{center}
\begin{tabular}{ll|l|cccccccccc}
\hline
                                              & \multicolumn{1}{c|}{}                   &                                                                                & \multicolumn{2}{c|}{Caltech256}                                               & \multicolumn{2}{c|}{Dogs}                                                     & \multicolumn{2}{c|}{Flowers}                                                  & \multicolumn{2}{c|}{CIFAR-10}                                                 & \multicolumn{2}{c}{CIFAR-100}                                                 \\
\multirow{-2}{*}{}                            & \multicolumn{1}{c|}{\multirow{-2}{*}{}} & \multirow{-2}{*}{\begin{tabular}[c]{@{}l@{}}Fine-tuning\\ method\end{tabular}} & NAcc                                  & \multicolumn{1}{c|}{RAcc}             & NAcc                                  & \multicolumn{1}{c|}{RAcc}             & NAcc                                  & \multicolumn{1}{c|}{RAcc}             & NAcc                                  & \multicolumn{1}{c|}{RAcc}             & NAcc                                  & RAcc                                  \\ \hline
\multicolumn{1}{l|}{}                         &                                         & Full param                                                                     & 81.12                                 & 64.52                                 & 69.11                                 & 39.28                                 & 85.35                                 & 68.37                                 & 97.74                                 & 83.99                                 & 79.19                                 & 54.56                                 \\
\multicolumn{1}{l|}{}                         &                                         & Linear probe                                                                   & 86.80                                 & 71.41                                 & 91.47                                 & 68.96                                 & 81.28                                 & 62.37                                 & 89.25                                 & 62.70                                 & 71.03                                 & 44.02                                 \\
\multicolumn{1}{l|}{}                         &                                         & BitFit (bias)                                                                  & 87.99                                 & 73.02                                 & 89.34                                 & 66.93                                 & 86.65                                 & 69.78                                 & 96.50                                 & 79.27                                 & 82.95                                 & 56.11                                 \\
\multicolumn{1}{l|}{}                         & \multirow{-4}{*}{PGD}                   & Adapter                                                                        & 89.37                                 & 74.76                                 & 89.36                                 & 67.28                                 & 83.53                                 & 66.16                                 & 97.50                                 & 82.55                                 & 84.21                                 & 57.47                                 \\ \cline{2-13} 
\multicolumn{1}{l|}{}                         &                                         & Full param                                                                     & 82.06                                 & 63.64                                 & 70.72                                 & 41.25                                 & 85.66                                 & 68.50                                 & 97.92                                 & 83.02                                 & 80.06                                 & 54.07                                 \\
\multicolumn{1}{l|}{}                         &                                         & Linear probe                                                                   & 86.83                                 & 71.30                                 & 91.71                                 & 68.99                                 & 81.04                                 & 62.53                                 & 89.53                                 & 61.99                                 & 71.27                                 & 43.70                                 \\
\multicolumn{1}{l|}{}                         &                                         & BitFit (bias)                                                                  & 88.63                                 & 72.85                                 & 89.91                                 & 66.55                                 & 86.44                                 & 69.05                                 & 96.81                                 & 78.46                                 & 83.46                                 & 55.00                                 \\
\multicolumn{1}{l|}{}                         &                                         & Adapter                                                                        & 89.61                                 & 74.52                                 & 90.16                                 & 66.96                                 & 82.78                                 & 65.00                                 & 97.70                                 & 81.66                                 & 84.50                                 & 56.72                                 \\ \cline{3-13} 
\multicolumn{1}{l|}{\multirow{-9}{*}{Swin}} & \multirow{-5}{*}{FGSM}                  & \textbf{Average diff.}                                                         & {\color[HTML]{32CB00} \textbf{+0.46}} & {\color[HTML]{CB0000} \textbf{-0.35}} & {\color[HTML]{32CB00} \textbf{+0.81}} & {\color[HTML]{32CB00} \textbf{+0.33}} & {\color[HTML]{CB0000} \textbf{-0.22}} & {\color[HTML]{CB0000} \textbf{-0.40}} & {\color[HTML]{32CB00} \textbf{+0.24}} & {\color[HTML]{CB0000} \textbf{-0.85}} & {\color[HTML]{32CB00} \textbf{+0.48}} & {\color[HTML]{CB0000} \textbf{-0.67}} \\ \hline
\end{tabular}
\end{center}
\vspace*{-3mm}
\caption{\label{tab:fgsm_pgd_performance_eps4} Performance of FGSM and PGD for $\varepsilon=4$ when fine-tuning a Swin transformer for multiple datasets. Individual performance is given for each fine-tuning method and dataset combination. The difference between FGSM and PGD averaged across fine-tuning methods is also reported for each dataset. Across all datasets combined, FGSM averages 0.39\% lower robustness and 0.35\% higher natural accuracy, which is closer to PGD than for $\varepsilon=8$.}
\vspace*{-3mm}
\end{table*}

\subsection{FGSM in robust transfer learning}

\textbf{PGD robustness.} We begin by evaluating the robustness of FGSM and PGD fine-tuned models against the PGD-10 attack (10 step PGD). Table~\ref{tab:fgsm_pgd_performance_eps8} reports the peak PGD-10 robustness (RAcc) and corresponding natural accuracy (NAcc) for adversarially fine-tuned models when $\varepsilon=8$. Individual results for each fine-tuning method and dataset are given along with the average NAcc and RAcc difference between FGSM and PGD. FGSM performs comparatively well against PGD, losing between 1.04-$2.08\%$ robustness while having between -0.06-$1.89\%$ higher natural accuracy on the Swin transformer (losing 1.46-$2.51\%$ robustness and gaining between 0.52-$1.61\%$ natural accuracy on the ViT model). There are a few outliers in performance. For example, for the Swin transformer, FGSM full fine-tuning on Caltech256 only loses 0.11\% robustness compared to PGD while having 1.54\% higher natural accuracy. On CIFAR-10, FGSM BitFit also only loses 0.42\% robustness while having 1.88\% higher natural accuracy.

Table~\ref{tab:fgsm_pgd_performance_eps4} reports the performance of PGD and FGSM for $\varepsilon=4$ on the Swin transformer. Compared to $\varepsilon=8$, models fine-tuned with FGSM are even closer in performance to PGD, with the absolute robustness gap being smaller for every dataset. On average, FGSM has between -0.33$-0.85\%$ lower robustness and -0.22$-0.81\%$ higher natural accuracy than PGD. These results follow the observations made in Section~\ref{sec:cosine}, where the adversarial examples of FGSM and PGD become more similar for smaller $\varepsilon$. 

For all settings of $\varepsilon=4$ and $\varepsilon=8$, \textit{FGSM remains stable without catastrophic overfitting}. Despite being a ``fast" method, FGSM robustness is surprisingly close to PGD for adversarial fine-tuning. However, this also does not consider the computational benefits of using FGSM.

\begin{table}[t]
\small
\begin{center}
\begin{tabular}{l|cc|cc|c}
\hline
           & \multicolumn{2}{c|}{Swin}        & \multicolumn{2}{c|}{ViT}         &  \multirow{2}{*}{\begin{tabular}[c]{@{}c@{}}$\frac{\text{FGSM}}{\text{PGD}}\%$\end{tabular}} \\ \cline{2-5}
           & \multicolumn{1}{c|}{PGD} & FGSM  & \multicolumn{1}{c|}{PGD} & FGSM  &                       \\ \hline
Caltech256 & 776.1                    & 192.3 & 670.3                    & 166.2 & \textbf{24.8\%}       \\
Dogs       & 379.9                    & 94.2  & 328.6                    & 81.5  & \textbf{24.8\%}       \\
Flowers    & 40.8                     & 10.3  & 35.3                     & 8.9   & \textbf{25.2\%}       \\
CIFAR-10   & 792.1                    & 195.4 &  684.2                   & 168.9 & \textbf{24.7\%}       \\
CIFAR-100  & 791.9                    & 195.4 &  684.4                   & 168.9 & \textbf{24.7\%}       \\ \hline
\end{tabular}
\end{center}
\vspace*{-3mm}
\caption{\label{tab:fgsm_pgd_train_time} Fine-tuning time (in minutes) for FGSM and PGD on different datasets, averaged across fine-tuning methods. $\frac{FGSM}{PGD}\%$ is the average percentage of time that FGSM uses relative to PGD. Across all datasets, FGSM averages $4.0\times$ less time than PGD-7.} 
\end{table}

\noindent\textbf{Computational cost.} We further compare the computational costs of using FGSM and PGD for adversarial fine-tuning. Table~\ref{tab:fgsm_pgd_train_time} reports the average time (in minutes) required for fine-tuning the Swin and ViT models on each dataset. In general, the trend is consistent between all models and datasets, with FGSM being roughly $4\times$ faster than PGD. The individual time for each fine-tuning method and dataset is also reported in the supplementary Table~\ref{tab:individual_timing}. Although not very significant in relation to the time added by PGD, PEFT methods help improve the computation time. For all datasets and models, full fine-tuning consistently takes the longest amount of time, while linear probing takes the least. BitFit and Adapter each take a similar amount of time, both of which are slightly less than full fine-tuning. 

\noindent\textbf{AutoAttack results.} To compare with stronger attacks, we use AutoAttack~\cite{croce2020reliable} to evaluate models trained on Stanford Dogs and Oxford Flowers102. We use the standard AutoAttack evaluation with APGD-CE, APGD-T, FAB-T~\cite{croce2020minimally}, and Square~\cite{andriushchenko2020square}.\footnote{We evaluate AutoAttack using a random subset of 1000 images from the test set. Once the images are selected, we use the same set of images for all models.} Table~\ref{tab:auto_attack} shows the AutoAttack robustness of PGD and FGSM models fine-tuned on the two datasets, with trends similar to PGD-10. For Stanford Dogs, the average difference in robustness between FGSM and PGD is 0.1\% for $\varepsilon=4$ and 0.5\% for $\epsilon=8$. For Oxford Flowers102, the difference is 0.5\% for $\varepsilon=4$ and 0.4\% for $\varepsilon=8$. Similarly to PGD adversarial fine-tuning, FGSM adversarial fine-tuning also yields models resistant to strong adversarial attacks like AutoAttack.

\subsection{Robust linear initialization (with FGSM)}

We extend our analysis of FGSM to the recently proposed Robust Linear Initialization (RoLI)~\cite{hua2024initialization}. For RoLI, a network is first initialized with a robust linear classification head prior to using another fine-tuning method (we use full fine-tuning here, but BitFit, Adapter, or other PEFTs will work). Although RoLI achieves high robustness, it requires two instances of adversarial fine-tuning: once with linear probing (to find the initialization of the linear head) and a second time with the additional fine-tuning method. PGD adversarial fine-tuning is expensive regardless of whether linear probing or another fine-tuning method is applied, and RoLI requires two instances of adversarial fine-tuning. 

We experiment with FGSM in RoLI. Since there are two instances of adversarial fine-tuning, there are four possible combinations of using FGSM or PGD. Table~\ref{tab:fgsm_roli} shows the PGD-10 robustness of RoLI on the Stanford Dogs dataset with $\varepsilon=4$ and $\varepsilon=8$. Compared to standard full fine-tuning on Stanford Dogs (from Tables~\ref{tab:fgsm_pgd_performance_eps8} and \ref{tab:fgsm_pgd_performance_eps4}), RoLI drastically improves the robustness of full fine-tuning by 23.64\% and 11.70\% on average for $\varepsilon=4$ and $\varepsilon=8$. 

Between the combinations of FGSM and PGD, there is no large difference in robustness. PGD-PGD (PGD linear probing and PGD full fine-tuning) only has 0.08\% and 1.22\% higher robustness than FGSM-FGSM for $\varepsilon=4$ and $\varepsilon=8$, despite requiring $4.2\times$ the amount of time. Surprisingly, for both $\varepsilon=4$ and $\varepsilon=8$, FGSM-PGD achieves the highest robustness, although it also takes roughly $2.6\times$ more time than FGSM-FGSM. In general, we conclude that there is no significant difference between using FGSM or PGD for the linear probing and initialization of RoLI. For the second stage of adversarial fine-tuning, PGD full fine-tuning results in a slightly higher robustness than FGSM, albeit at a higher computational cost.

\begin{table}[t]
\begin{center}
\begin{tabular}{lcccc}
\hline
\multicolumn{5}{c}{Dogs}                                                                                                         \\ \hline
\multicolumn{1}{l|}{}                 & \multicolumn{2}{c|}{$\varepsilon=4$}               & \multicolumn{2}{c}{$\varepsilon=8$} \\
\multicolumn{1}{l|}{}                 & PGD           & \multicolumn{1}{c|}{FGSM}          & PGD              & FGSM             \\ \hline
\multicolumn{1}{l|}{Full param.}      & 32.4          & \multicolumn{1}{c|}{33.3}          & 20.2             & 20.8             \\
\multicolumn{1}{l|}{Linear probe}     & 52.8          & \multicolumn{1}{c|}{54.3}          & 26.3             & 25.2             \\
\multicolumn{1}{l|}{BitFit (bias)}    & 56.9          & \multicolumn{1}{c|}{55.4}          & 29.8             & 29.0             \\
\multicolumn{1}{l|}{Adapter}          & 56.6          & \multicolumn{1}{c|}{55.5}          & 29.2             & 28.7             \\ \hline
\multicolumn{1}{l|}{\textbf{Average}} & \textbf{49.7} & \multicolumn{1}{c|}{\textbf{49.6}} & \textbf{26.4}    & \textbf{25.9}    \\ \hline
\multicolumn{5}{c}{Flowers}                                                                                                      \\ \hline
\multicolumn{1}{l|}{}                 & \multicolumn{2}{c|}{$\varepsilon=4$}               & \multicolumn{2}{c}{$\varepsilon=8$} \\
\multicolumn{1}{l|}{}                 & PGD           & \multicolumn{1}{c|}{FGSM}          & PGD              & FGSM             \\ \hline
\multicolumn{1}{l|}{Full param.}      & 64.6          & \multicolumn{1}{c|}{64.9}          & 43.1             & 43.7             \\
\multicolumn{1}{l|}{Linear probe}     & 49.3          & \multicolumn{1}{c|}{47.9}          & 32.8             & 33.0             \\
\multicolumn{1}{l|}{BitFit (bias)}    & 63.4          & \multicolumn{1}{c|}{61.8}          & 45.6             & 42.9             \\
\multicolumn{1}{l|}{Adapter}          & 57.0          & \multicolumn{1}{c|}{57.8}          & 44.6             & 45.0             \\ \hline
\multicolumn{1}{l|}{\textbf{Average}} & \textbf{58.6} & \multicolumn{1}{c|}{\textbf{58.1}} & \textbf{41.5}    & \textbf{41.1}    \\ \hline
\end{tabular}
\end{center}
\vspace*{-5mm}
\caption{\label{tab:auto_attack} AutoAttack~\cite{croce2020reliable} robustness for FGSM and PGD models trained on Stanford Dogs and Oxford Flowers 102 for $\varepsilon=4$ and 8. FGSM performs close to PGD for both datasets. For Dogs, the average gap is 0.1\% and 0.5\% for $\varepsilon=4$ and $\varepsilon=8$. For flowers, the average gap is 0.5\% and 0.4\% instead.}
\vspace*{-1mm}
\end{table}

\begin{table}[t]
\begin{center}
\begin{tabular}{llccc}
\hline
\multicolumn{5}{c}{$\varepsilon = 4$}                                                                                                                                                                                                                    \\ \hline
\multicolumn{1}{l|}{\begin{tabular}[c]{@{}l@{}}Robust lin.\\ probing\end{tabular}} & \multicolumn{1}{l|}{\begin{tabular}[c]{@{}l@{}}Additional\\ fine-tuning\end{tabular}} & NAcc  & RAcc  & \begin{tabular}[c]{@{}c@{}}Total \\ time (min)\end{tabular} \\ \hline
\multicolumn{1}{l|}{PGD}                                                           & \multicolumn{1}{l|}{PGD}                                                              & 89.39 & 64.64 & 749.0                                                       \\
\multicolumn{1}{l|}{PGD}                                                           & \multicolumn{1}{l|}{FGSM}                                                             & 89.32 & 64.63 & 465.1                                                       \\
\multicolumn{1}{l|}{FGSM}                                                          & \multicolumn{1}{l|}{PGD}                                                              & 90.02 & 65.73 & 465.3                                                       \\
\multicolumn{1}{l|}{FGSM}                                                          & \multicolumn{1}{l|}{FGSM}                                                             & 89.50 & 64.56 & 181.8                                                       \\ \hline
\multicolumn{5}{c}{$\varepsilon = 8$}                                                                                                                                                                                                                    \\ \hline
\multicolumn{1}{l|}{\begin{tabular}[c]{@{}l@{}}Robust lin.\\ probing\end{tabular}} & \multicolumn{1}{l|}{\begin{tabular}[c]{@{}l@{}}Additional\\ fine-tuning\end{tabular}} & NAcc  & RAcc  & \begin{tabular}[c]{@{}c@{}}Total \\ time (min)\end{tabular} \\ \hline
\multicolumn{1}{l|}{PGD}                                                           & \multicolumn{1}{l|}{PGD}                                                              & 86.15 & 36.56 & 749.2                                                       \\
\multicolumn{1}{l|}{PGD}                                                           & \multicolumn{1}{l|}{FGSM}                                                             & 87.06 & 35.47 & 465.2                                                       \\
\multicolumn{1}{l|}{FGSM}                                                          & \multicolumn{1}{l|}{PGD}                                                              & 86.38 & 37.35 & 466.1                                                       \\
\multicolumn{1}{l|}{FGSM}                                                          & \multicolumn{1}{l|}{FGSM}                                                             & 88.11 & 35.34 & 181.5                                                       \\ \hline
\end{tabular}
\end{center}
\vspace*{-3mm}
\caption{\label{tab:fgsm_roli} Robust Linear Initialization (RoLI)~\cite{hua2024initialization} with FGSM or PGD on the Stanford Dogs dataset. RoLI is applied in two stages, once with linear probing and an additional time with a different method (full fine-tuning here). Natural accuracy (NAcc), PGD-10 robustness (RAcc), and total fine-tuning time is reported for each combination of using FGSM or PGD for $\varepsilon=4$ and $\varepsilon=8$.} 
\vspace*{-1mm}
\end{table}
\section{Discussion}

FGSM has always been known to suffer from catastrophic overfitting~\cite{Wong2020Fast}. However, one of the key outcomes of this work is finding the inherent stability of FGSM in robust transfer learning. Although much prior work has been done to mitigate catastrophic overfitting in fast adversarial training~\cite{shafahi2019adversarial, shafahi2019adversarial, andriushchenko2020understanding, jia2022prior}, this is much less useful for robust transfer learning. Instead, robust transfer learning offers an alternative setting where stability may not be a main concern. As a result, regularization in fast adversarial fine-tuning can be shifted to improving the FGSM adversarial example estimation in order to make the performance gap with PGD smaller. 

Efficient robust transfer learning also offers other directions for future work. For example, \cite{Shafahi2020Adversarially} discusses how the similarity of the pre-training and downstream datasets plays a role in how much robustness can be inherited in the downstream task. Factors such as image dimension, number of classes, or pre-training $\varepsilon$ can all play a role in the final performance. An interesting direction for future work can be to explore how different transfer learning scenarios and pre-trained model settings can impact the estimation difference between the adversarial examples of FGSM and PGD.

\section{Conclusions}

In this paper, we investigate the use of FGSM for efficient adversarial fine-tuning in robust transfer learning. We first find that FGSM is much more stable to catastrophic overfitting during fine-tuning than when training from scratch. With parameter-efficient fine-tuning methods, FGSM becomes even more stable, where models fine-tuned with linear probing can avoid catastrophic overfitting even up to $\varepsilon=32$. We demonstrate that this stability allows FGSM to perform well without relying on additional regularization, and we further validate FGSM's fine-tuning performance on multiple datasets. For $\varepsilon=4$ and $\varepsilon=8$, FGSM uses $4\times$ less training time than PGD while only losing 0.39\% and 1.39\% test robustness on average (and having 0.35\% and 1.21\% higher natural accuracy). Surprisingly, for robust transfer learning, FGSM can act not only as an efficient alternative to PGD, but also as a well-performing one. 






{
    \small
    \bibliographystyle{ieeenat_fullname}
    \bibliography{main}
}

\clearpage
\setcounter{page}{1}
\maketitlesupplementary

\section{Experimental details}
\label{sec:suppl_exp_details}

We use the Swin-B and ViT-B/16 models from ARES 2.0~\cite{dong2020benchmarking} in this paper. The models are trained om ImageNet-1K~\cite{deng2009imagenet} with $\varepsilon=4$. In order to ensure that the downstream dataset images match the pre-training model specifications, images are also resized to $224\times224$. For the optimizer hyperparameters, we use the SGD optimizer with learning rate 0.05 (for RoLI full fine-tuning, our initial learning rate is 0.005), momentum 0.9, and weight decay 5e-4 by default. The learning rate is decreased by a factor of 0.1 at $\frac{1}{4}$ and $\frac{3}{4}$ the total epochs. For each experimental setting, we additionally tune the learning rate in order to ensure that fine-tuned models reach convergence. Given the large number of experiments and different settings in the work, we do not optimize all hyperparameters for each setting. For Adapter~\cite{houlsby2019parameter}, a reduction factor of 8 is used.

A training batch size of 128 is used. By default, the Caltech256~\cite{griffin2007caltech} and Stanford Dogs~\cite{khosla2011novel} datasets are fine-tuned for 40 epochs and the Oxford Flowers102~\cite{nilsback2008automated}, CIFAR-10 and CIFAR-100~\cite{krizhevsky2009learning} datasets are fine-tuned for 20 epochs. The only exception was with $\varepsilon=8$ on a Swin transfer using Adapter for the Flowers dataset, as 40 epochs were required for both PGD and FGSM to converge to a good solution. We hypothesize that this due to the small training set of Flowers (2040 training images). Since Adapter adds completely new layers, more data may be required compared to the other methods.

For the FGSM variants in Section~\ref{sec:FGSM_variants}, we apply the same settings as in the original work. For the random initialization of FGSM+RI~\cite{Wong2020Fast}, we use $\delta=Uniform(-\varepsilon,\varepsilon)$ along with a step size of $\alpha = 1.25\varepsilon$. For GradAlign~\cite{andriushchenko2020understanding}, we use the gradient alignment regularizer of $1-Cos(\nabla_x(x,y,\theta),\nabla_x(x+\delta))$ along with a regularization coefficient of 0.2.

\section{Additional results}
\label{sec:additional_results}


We include some additional experimental results along with individual numbers here.

\begin{figure}[t]
\vspace*{-0mm}
    \centering
   \includegraphics[width=0.8\linewidth,trim={0mm 0mm 0mm 0mm},clip]{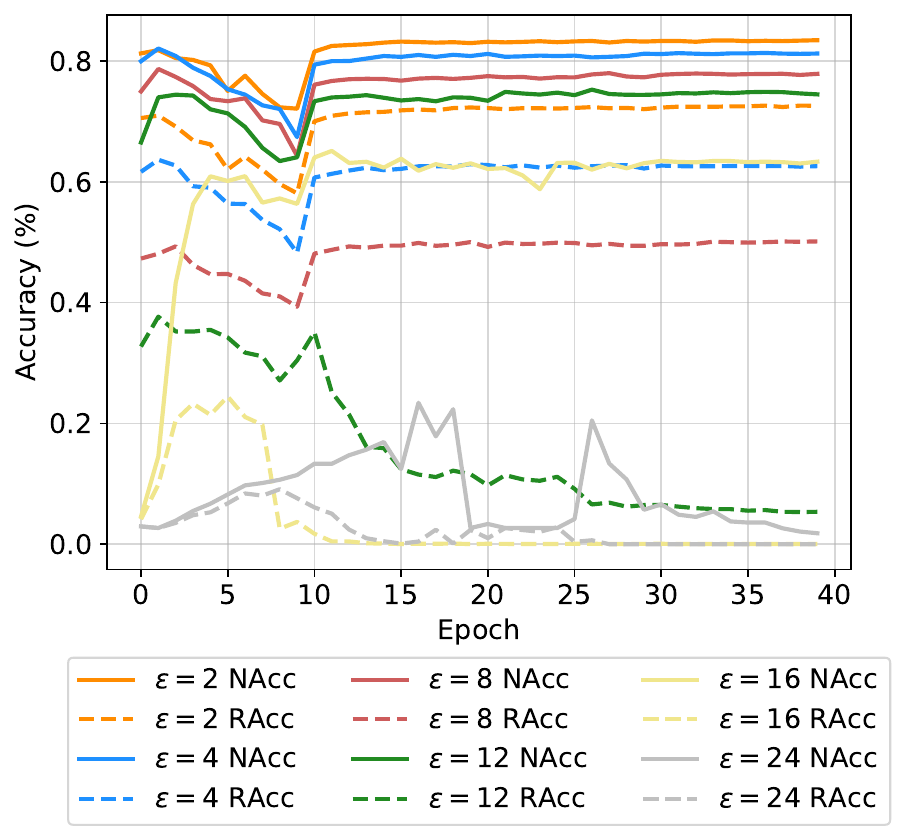}
  \caption{Natural accuracy (solid line) and robustness (dashed line) for FGSM full fine-tuning when trained with different epsilon values. When $\varepsilon<12$, model training is consistent without catastrophic overfitting. At $\varepsilon\geq12$, the model begins to catastrophically overfit (although the best performing model of $\varepsilon=12$ performs close to PGD).}
  \label{fig:fgsm_fft_vary_eps} 
\vspace*{-3mm}
\end{figure}

\begin{figure}[t]
\vspace*{-0mm}
  \centering
  \begin{subfigure}{1.0\linewidth}
    \includegraphics[width=0.8\linewidth,trim={0mm 0mm 0mm 0mm},clip]{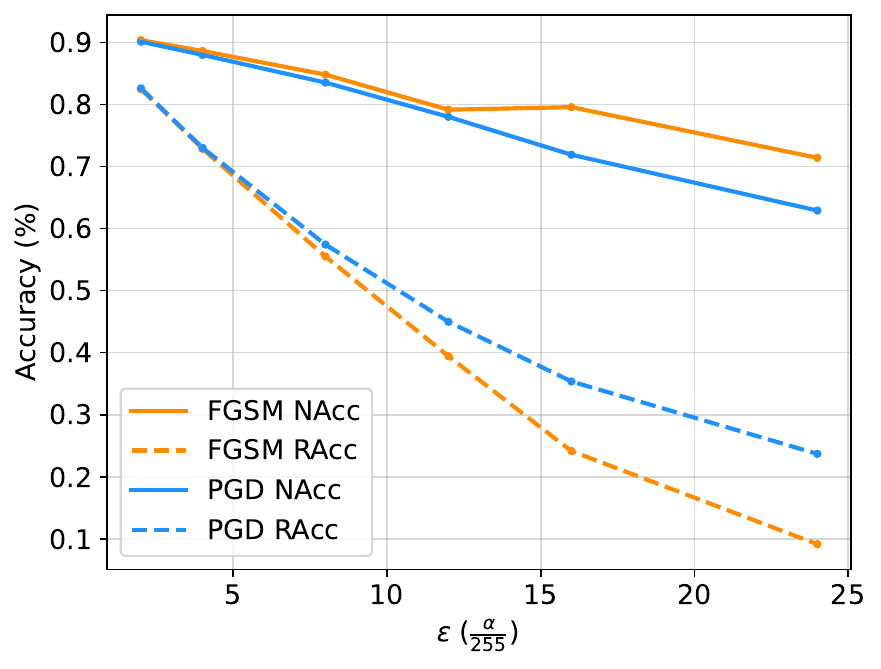}
    \caption{BitFit (bias)}
    \label{fig:fgsm_pgd_bitfit_veps}
  \end{subfigure}
  \vspace*{3mm}
    \begin{subfigure}{1.0\linewidth}
    \includegraphics[width=0.8\linewidth,trim={0mm 0mm 0mm 0mm},clip]{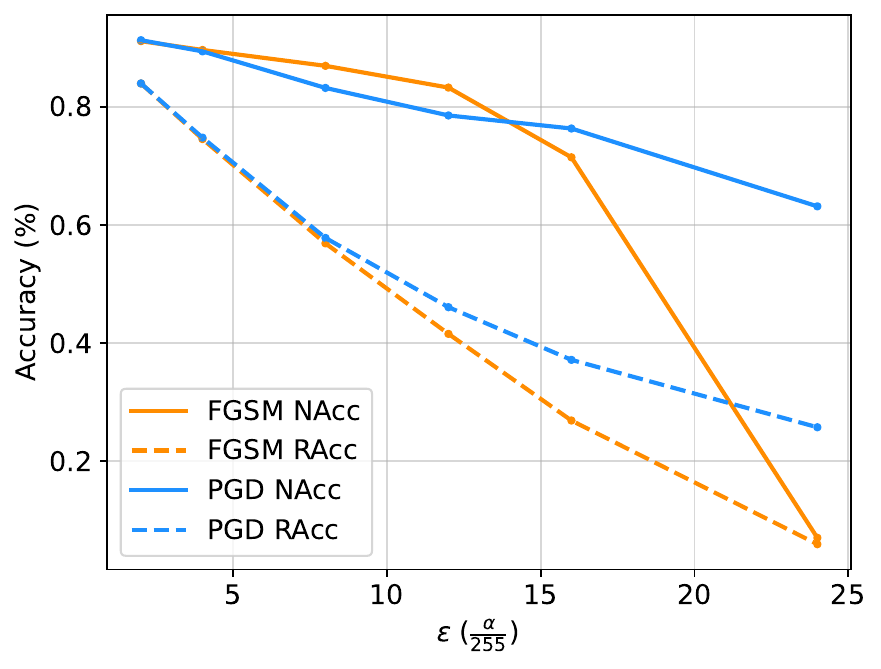}
    \caption{Adapter}
    \label{fig:fgsm_pgd_adapter_veps}
  \end{subfigure}
  \vspace*{-8mm}
  \caption{\label{fig:vary-epsilon-bitfit-adapter} FGSM vs. PGD (a) BitFit (bias) and (b) adapter natural accuracy (NAcc) and robust accuracy (RAcc) for varying $\varepsilon$ values on Caltech256 with a Swin transformer. Both methods remain stable and achieve reasonable performance up to $\varepsilon=12$. At $\varepsilon=16$, the gap between PGD and FGSM becomes much larger for both BitFit (bias) and adapter fine-tuning (where BitFit does not catastrophically overfit, while adapter does).}
  \vspace*{-1mm}
\end{figure}

\begin{figure}[t]
\vspace*{-0mm}
    \centering
   \includegraphics[width=0.8\linewidth,trim={0mm 0mm 0mm 0mm},clip]{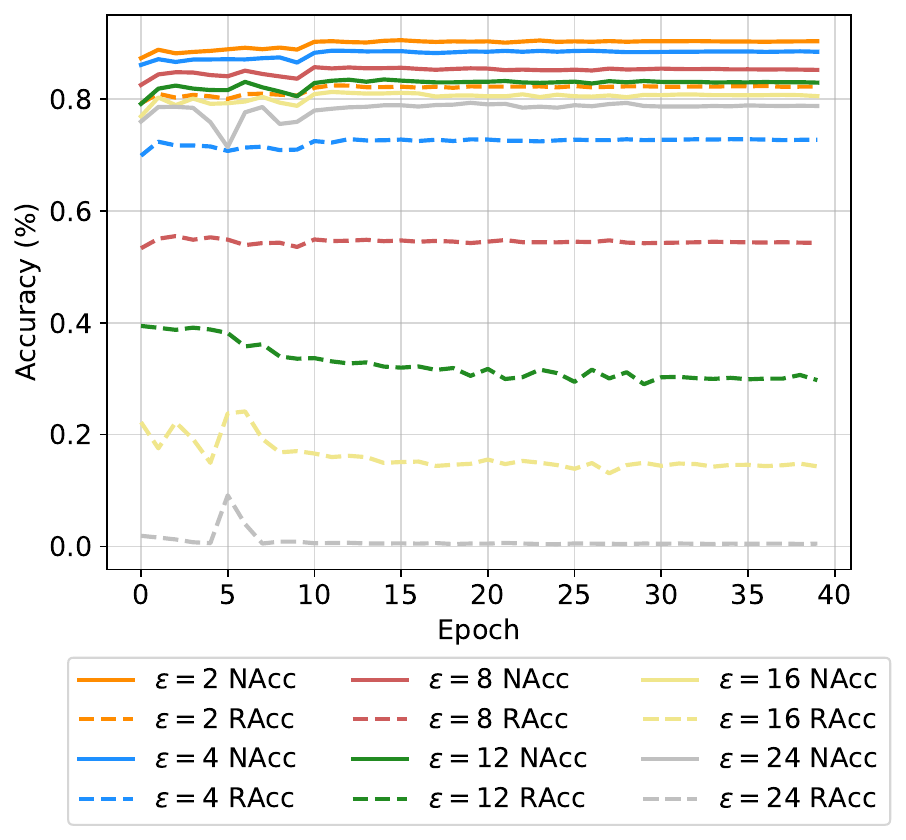}
   \vspace*{-2mm}
  \caption{Natural accuracy (solid line) and robustness (dashed line) for FGSM BitFit (bias) fine-tuning when trained with different epsilon values. When $\varepsilon<24$, model training is relatively consistent without catastrophic overfitting. At $\varepsilon=24$, the model generally cannot converge to a good solution anymore.}
  \label{fig:fgsm_bitfit_vary_eps} 
\vspace*{-3mm}
\end{figure}

\begin{figure}[t]
\vspace*{-0mm}
    \centering
   \includegraphics[width=0.8\linewidth,trim={0mm 0mm 0mm 0mm},clip]{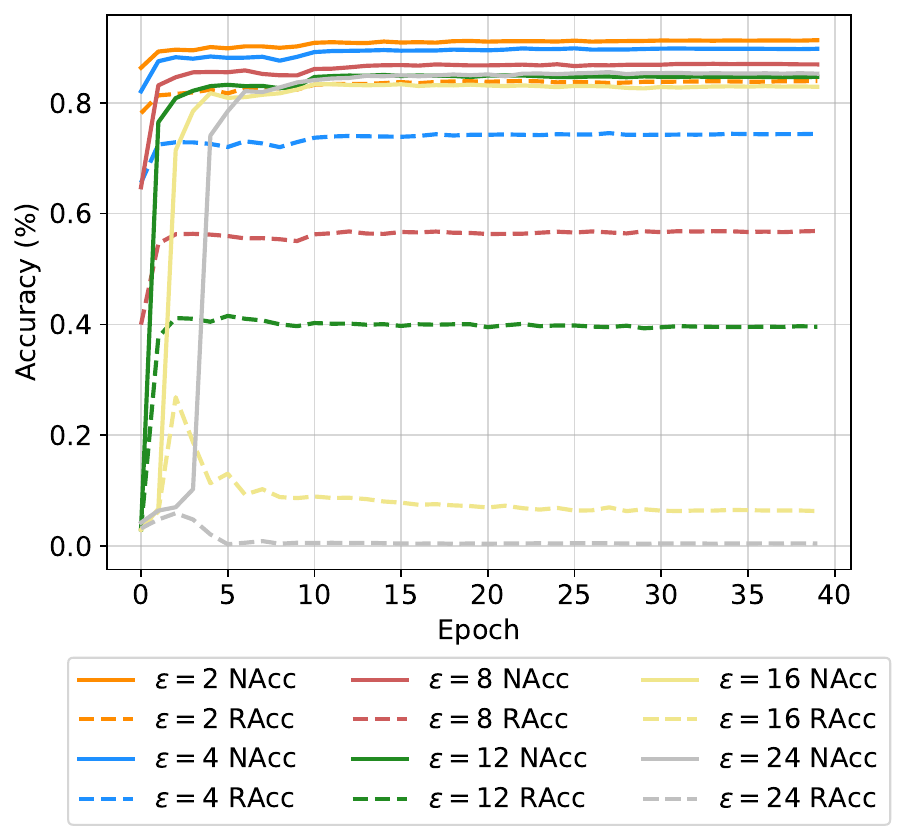}
   \vspace*{-2mm}
  \caption{Natural accuracy (solid line) and robustness (dashed line) for FGSM adapter fine-tuning when trained with different epsilon values. When $\varepsilon<16$, model training is consistent without catastrophic overfitting. At $\varepsilon\geq16$, the model begins to catastrophically overfit.}
  \label{fig:fgsm_adapter_vary_eps} 
\vspace*{-3mm}
\end{figure}

\noindent\textbf{Varying epsilon.} Figure~\ref{fig:fgsm_fft_vary_eps} shows the natural accuracy and robustness curves for FGSM full fine-tuning on Caltech256 with multiple $\varepsilon$. Catastrophic overfitting begins at $\varepsilon=12$, although the peak robustness difference between PGD and FGSM is not too large (3.23\%). At $\varepsilon\geq16$, catastrophic overfitting becomes a bigger problem for performance, and the gap increases. 

Figures~\ref{fig:fgsm_pgd_bitfit_veps} and \ref{fig:fgsm_pgd_adapter_veps} plot the PGD and FGSM natural accuracy (NAcc) and PGD-10 robustness (RAcc) for varying $\varepsilon$ with BitFit and Adapter. Figures~\ref{fig:fgsm_bitfit_vary_eps} and \ref{fig:fgsm_adapter_vary_eps} show the natural accuracy and robustness curves of FGSM for each. Both BitFit and Adapter remain stable up through $\varepsilon=12$. At $\varepsilon=16$, BitFit does not overfit, but Adapter begins to overfit. At $\varepsilon=24$, BitFit overfits.

Tables~\ref{tab:FFT_vary_eps}, \ref{tab:LP_vary_eps}, \ref{tab:bitfit_vary_eps}, and \ref{tab:adapter_vary_eps} report the natural accuracy and robustness of the model state with the highest robustness for each $\varepsilon$ for full (paramter) fine-tuning, linear probing, BitFit, and Adapter. In general, FGSM and PGD have relatively close performance when $\varepsilon\leq12$.

\begin{table*}[]
\begin{center}
\begin{tabular}{ll|l|ccccc}
\hline
                                           & \multicolumn{1}{c|}{} & \multicolumn{1}{c|}{\begin{tabular}[c]{@{}c@{}}Fine-tuning\\ method\end{tabular}} & Caltech256 & Dogs  & Flowers & CIFAR-10 & CIFAR-100 \\ \hline
\multicolumn{1}{l|}{\multirow{8}{*}{Swin}} & \multirow{4}{*}{PGD}  & Full param                                                                        & 810.3      & 396.7 & 33.9    & 827.5    & 827.3     \\
\multicolumn{1}{l|}{}                      &                       & Linear probe                                                                      & 718.9      & 352.0 & 30.2    & 733.4    & 733.7     \\
\multicolumn{1}{l|}{}                      &                       & BitFit (bias)                                                                     & 779.0      & 381.1 & 32.6    & 794.2    & 794.6     \\
\multicolumn{1}{l|}{}                      &                       & Adapter                                                                           & 796.4      & 389.9 & 66.6    & 813.2    & 812.1     \\ \cline{2-8} 
\multicolumn{1}{l|}{}                      & \multirow{4}{*}{FGSM}                  & Full param                                                                        & 231.4      & 113.3 & 9.8     & 235.4    & 235.5     \\
\multicolumn{1}{l|}{}                      &                       & Linear probe                                                                      & 139.8      & 68.7  & 6.1     & 141.9    & 141.9     \\
\multicolumn{1}{l|}{}                      &                       & BitFit (bias)                                                                     & 199.6      & 97.7  & 8.5     & 202.8    & 202.7     \\
\multicolumn{1}{l|}{}                      &                       & Adapter                                                                           & 198.3      & 97.2  & 16.8    & 201.6    & 201.7     \\ \hline
\multicolumn{1}{l|}{\multirow{8}{*}{ViT}}  & \multirow{4}{*}{PGD}  & Full param                                                                        & 702.8      & 344.4 & 29.3    & 717.8    & 716.9     \\
\multicolumn{1}{l|}{}                      &                       & Linear probe                                                                      & 625.2      & 306.8 & 26.3   & 638.1    & 638.7     \\
\multicolumn{1}{l|}{}                      &                       & BitFit (bias)                                                                     & 669.7      & 328.4 & 28.0    & 683.3    & 684.2     \\
\multicolumn{1}{l|}{}                      &                       & Adapter                                                                           & 683.5      & 334.9 & 57.2    & 697.6    & 698.1     \\ \cline{2-8} 
\multicolumn{1}{l|}{}                      & \multirow{4}{*}{FGSM} & Full param                                                                        & 202.1      & 98.9  & 8.5     & 205.7    & 205.3     \\
\multicolumn{1}{l|}{}                      &                       & Linear probe                                                                      & 124.1      & 61.0  & 5.4     & 126.1    & 126.0     \\
\multicolumn{1}{l|}{}                      &                       & BitFit (bias)                                                                     & 169.0      & 82.8  & 7.2     & 171.7    & 172.0     \\
\multicolumn{1}{l|}{}                      &                       & Adapter                                                                           & 169.6      & 83.1  & 14.5    & 172.1    & 172.2     \\ \hline
\end{tabular}
\end{center}
\vspace*{-3mm}
\caption{\label{tab:individual_timing} Individual timing table (in minutes) for each fine-tuning method and dataset.}
\vspace*{-5mm}
\end{table*}

\begin{table}[t]
\begin{center}
\begin{tabular}{l|cc|cc}
\hline
                   & \multicolumn{2}{c|}{PGD} & \multicolumn{2}{c}{FGSM} \\
                   & NAcc        & RAcc       & NAcc        & RAcc       \\ \hline
$\varepsilon = 2$  & 82.95       & 72.21      & 83.34       & 72.61      \\
$\varepsilon = 4$  & 81.12       & 64.52      & 82.06       & 63.64      \\
$\varepsilon = 8$  & 76.33       & 50.42      & 77.87       & 50.13      \\
$\varepsilon = 12$ & 71.58       & 40.90      & 73.98       & 37.67      \\
$\varepsilon = 16$ & 66.38       & 34.45      & 60.13       & 24.45      \\
$\varepsilon = 24$ & 2.94        & 2.94       & 10.67       & 9.10       \\ \hline
\end{tabular}
\vspace*{-2mm}
\end{center}
\caption{\label{tab:FFT_vary_eps} Results for full (parameter) fine-tuning on Caltech256 for varying $\varepsilon$. Natural accuracy (NAcc) and PGD-10 robustness (RAcc) reported for the model with peak robustness at each $\varepsilon$.}
\vspace*{-3mm}
\end{table}

\begin{table}[t]
\begin{center}
\begin{tabular}{l|cc|cc}
\hline
                   & \multicolumn{2}{c|}{PGD} & \multicolumn{2}{c}{FGSM} \\
                   & NAcc        & RAcc       & NAcc        & RAcc       \\ \hline
$\varepsilon = 2$  & 87.14       & 80.17      & 87.14       & 80.17      \\
$\varepsilon = 4$  & 86.80       & 71.41      & 86.83       & 71.30      \\
$\varepsilon = 8$  & 85.76       & 53.43      & 86.65       & 52.30      \\
$\varepsilon = 16$ & 78.85       & 31.07      & 81.67       & 27.23      \\
$\varepsilon = 24$ & 80.41       & 17.84      & 80.43       & 11.94      \\
$\varepsilon = 32$ & 71.07       & 10.88      & 79.30       & 5.01       \\ \hline
\end{tabular}
\vspace*{-2mm}
\end{center}
\caption{\label{tab:LP_vary_eps} Results for linear probing on Caltech256 for varying $\varepsilon$. Natural accuracy (NAcc) and PGD-10 robustness (RAcc) reported for the model with peak robustness at each $\varepsilon$.}
\vspace*{-3mm}
\end{table}

\begin{table}[t]
\small
\begin{center}
\begin{tabular}{l|cc|cc}
\hline
                   & \multicolumn{2}{c|}{PGD} & \multicolumn{2}{c}{FGSM} \\
                   & NAcc        & RAcc       & NAcc        & RAcc       \\ \hline
$\varepsilon = 2$  & 90.12       & 82.60      & 90.36       & 82.46      \\
$\varepsilon = 4$  & 87.99       & 73.02      & 88.63       & 72.85      \\
$\varepsilon = 8$  & 83.53       & 57.42      & 84.81       & 55.50      \\
$\varepsilon = 12$ & 78.01       & 44.97      & 79.16       & 39.43      \\
$\varepsilon = 16$ & 71.90       & 35.36      & 79.57       & 24.14      \\
$\varepsilon = 24$ & 62.90       & 23.69      & 71.41       & 9.13       \\ \hline
\end{tabular}
\vspace*{-2mm}
\end{center}
\caption{\label{tab:bitfit_vary_eps} Results for BitFit (bias) on Caltech256 for varying $\varepsilon$. Natural accuracy (NAcc) and PGD-10 robustness (RAcc) reported for the model with peak robustness at each $\varepsilon$.}
\vspace*{-3mm}
\end{table}

\begin{table}[t]
\begin{center}
\begin{tabular}{l|cc|cc}
\hline
                   & \multicolumn{2}{c|}{PGD} & \multicolumn{2}{c}{FGSM} \\
                   & NAcc        & RAcc       & NAcc        & RAcc       \\ \hline
$\varepsilon = 2$  & 91.26       & 83.94      & 91.10       & 83.93      \\
$\varepsilon = 4$  & 89.37       & 74.76      & 89.61       & 74.52      \\
$\varepsilon = 8$  & 83.18       & 57.78      & 86.92       & 56.86      \\
$\varepsilon = 12$ & 78.52       & 46.06      & 83.24       & 41.54      \\
$\varepsilon = 16$ & 76.31       & 37.14      & 71.43       & 26.87      \\
$\varepsilon = 24$ & 63.13       & 25.73      & 7.01        & 5.93       \\ \hline
\end{tabular}
\vspace*{-3mm}
\end{center}
\caption{\label{tab:adapter_vary_eps} Results for Adapter on Caltech256 for varying $\varepsilon$. Natural accuracy (NAcc) and PGD-10 robustness (RAcc) reported for the model with peak robustness at each $\varepsilon$.}
\vspace*{-3mm}
\end{table}

\noindent\textbf{Individual timing.} Table~\ref{tab:individual_timing} reports the individual time required for each fine-tuning method on each dataset for PGD and FGSM. In general, PEFT methods offer a small improvement in computation time. Full (parameter) fine-tuning consistently takes the longest time, while linear probing takes the least. BitFit and Adapter each take less time than full fine-tuning, although more time than linear probing. Linear probing offers a 39\% time improvement compared to full fine-tuning for FGSM, but only an 11\% improvement for PGD. Since the absolute amount of time decreased by linear probing is roughly the same for both PGD and FGSM, this once again highlights how much overhead PGD adversarial fine-tuning adds.

\begin{figure}[t]
\vspace*{-2mm}
    \centering
   \includegraphics[width=0.75\linewidth,trim={0mm 0mm 0mm 0mm},clip]{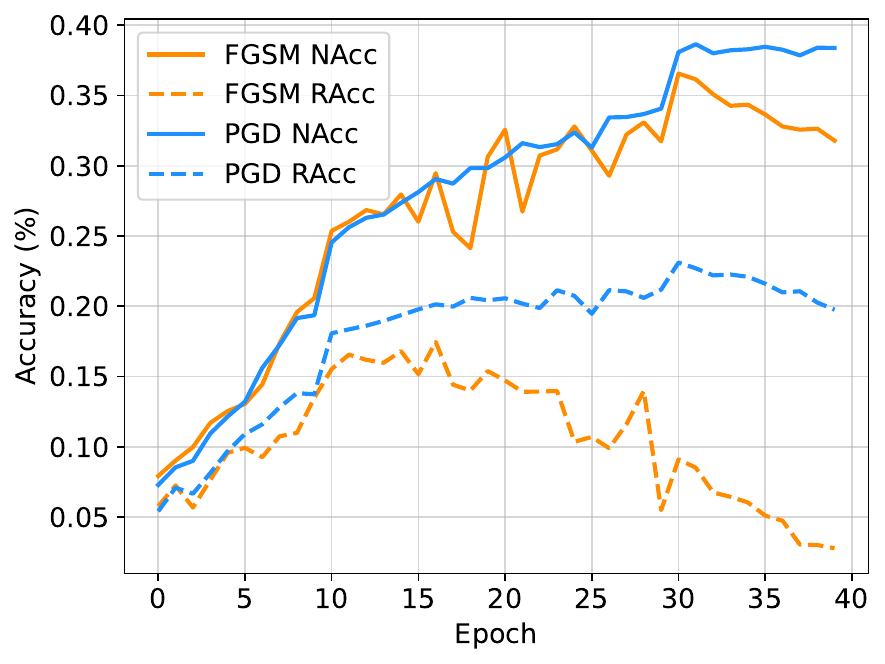}
  \caption{Adversarial training with FGSM and PGD from scratch on Caltech256 on a ResNet-50 with $\varepsilon=8$. Contrary to adversarial transfer learning at $\varepsilon=8$, FGSM training overfits at the end and achieves a near zero PGD test robustness.}
  \vspace*{-2mm}
  \label{fig:fgsm_pgd_c256_scratch} 
\vspace*{-0mm}
\end{figure}

\noindent\textbf{Adversarially training a model from scratch.} Figure~\ref{fig:fgsm_pgd_c256_scratch} shows the natural accuracy and robustness curves for a ResNet-50~\cite{he2016deep} adversarially trained from scratch using PGD and FGSM (we use a ResNet-50 here due to the much higher cost of adversarially training a transformer from scratch (data, time required to converge, and the cost of tuning hyperparameters. When adversarially training a transformer from scratch on CIFAR-10, even after 40 epochs, the training accuracy is barely at 26\% and increasing slowly). The dataset used is Caltech256 with $\varepsilon=8$. The blue lines report the performance for the PGD trained model and the orange lines show the performance of the FGSM model. Overall, the PGD model maintains a solid PGD-10 testing robustness (RAcc) while the FGSM trained model begins to overfit. By the end of training, the FGSM adversarially trained model has a near-zero PGD-10 testing robustness. In general, when training from scratch, catastrophic overfitting serves as a large challenge for FGSM regardless of model architecture or dataset~\cite{kim2021understanding, wu2022towards, shao2021adversarial, gopal2025safer}.


\end{document}